\documentclass[11pt]{article}
\usepackage[hyperref]{acl2023}
\usepackage{times}
\usepackage{latexsym}
\usepackage{amsmath}
\usepackage{amssymb}
\usepackage{booktabs}
\usepackage{multirow}
\usepackage{graphicx}
\usepackage{xcolor}
\usepackage{microtype}
\usepackage{inconsolata}
\usepackage{url}
\usepackage{array}

\definecolor{passgreen}{rgb}{0.13,0.55,0.13}
\definecolor{failred}{rgb}{0.80,0.10,0.10}
\definecolor{warnorg}{rgb}{0.85,0.40,0.00}

\newcommand{\best}[1]{\textbf{#1}}

\title{It's Not the Capability: Harness Sensitivity Is\\Non-Monotone Across LLM Agent Tiers}

\author{
  Yong-eun Cho \\
  KailosLab \\
  Seoul, Republic of Korea \\
  \texttt{kevin@kailoslab.com}
}

\begin{document}
\maketitle

\begin{abstract}
A prevalent assumption in LLM agent deployment holds that more structured
harnesses universally improve reliability, and that higher-capability models
need proportionally less structural guidance---together implying a
\emph{monotone inverse} relationship between model capability tier and
optimal harness complexity.
We test this hypothesis through a controlled 432-run experiment crossing
six models across four capability tiers with three harness conditions
(\textit{light}, \textit{balanced}, \textit{strict}) on HEAT-24, a
24-task synthetic benchmark with git-based workspace verification.
Our results refute the monotone inverse relationship on two fronts.
First, for the frontier \emph{chat} model evaluated (Gemini 2.5 Flash),
increased harness verbosity \emph{lowers} VTSR by 29--38 percentage
points---a \textbf{harness-complexity paradox}.
Second, for the frontier \emph{reasoning} model evaluated (Qwen3.5-122B,
extended thinking enabled), strict harness achieves the \emph{highest} VTSR
(91.7\%) and the \emph{lowest} latency, the opposite of the prediction.
Within the constrained tier, a 2\,B model (Gemma4:e2B) matches
strong-open-tier stability at 91.7\% across all harnesses.
Because each tier is represented by a single model in this study,
these results should be interpreted as model-specific observations;
harness sensitivity appears \textbf{non-monotone} across the models
evaluated, and depends critically on model type (chat vs.\ reasoning).
We introduce a six-label failure taxonomy showing that
\texttt{format\_violation} dominates capable-model failures while
\texttt{wrong\_file} dominates low-capability failures, and we derive
practical tier-aware harness selection guidelines.
\end{abstract}

\section{Introduction}
\label{sec:intro}

Autonomous LLM agents that read, reason about, and modify workspace artifacts
are increasingly deployed in software engineering~\citep{liu2023agentbench,
jimenez2024swebench}, document processing, and operational workflows.
The quality of the \emph{harness}---the system-level prompt that specifies
task scope, allowed operations, output format, and verification procedure---is
widely believed to be a primary lever for improving agent reliability
\citep{yao2022react, shinn2023reflexion}.

Existing benchmarks evaluate agents on a fixed harness and report aggregate
accuracy, obscuring the interaction between harness complexity and model
capability.
Practitioners routinely apply strict, highly-structured harnesses to all models
in a deployment fleet under two implicit assumptions: that more structure always
improves reliability, and that higher-capability models need less structural
guidance---forming a \emph{monotone inverse} relationship between capability
tier and optimal harness complexity.
We ask whether this monotone inverse hypothesis holds empirically across
a diverse set of capability tiers, and whether the answer differs between
chat-oriented and reasoning-oriented frontier models.

We make the following contributions:
\begin{itemize}
  \item \textbf{HEAT-24} (\textbf{H}arness \textbf{E}valuation for
        \textbf{A}gent \textbf{T}asks), a deterministic 24-task synthetic
        benchmark with workspace-level git-based verification covering six
        task categories.
  \item The first controlled empirical test of the monotone inverse
        capability-harness hypothesis, crossing six models across four
        capability tiers with three harness conditions (432 total runs).
  \item Evidence that the hypothesis fails in opposite directions simultaneously:
        a \textbf{harness-complexity paradox} (strict harness \emph{hurts}
        the frontier chat model) and a \textbf{non-monotonic pattern} (strict
        harness \emph{helps} the frontier reasoning model most).
  \item The finding that parameter count is an unreliable proxy for
        harness sensitivity: a 2\,B model (Gemma4:e2B) matches strong-open-tier
        stability, demonstrating that instruction-tuning quality is the true
        moderating variable.
  \item A six-label failure taxonomy and practical \textbf{tier-aware and
        type-aware harness selection} guidelines.
\end{itemize}

\section{Related Work}
\label{sec:related}

\paragraph{LLM agent benchmarks.}
\citet{liu2023agentbench} evaluate LLMs across eight interactive environments
and show strong capability gaps between frontier and open-source models.
\citet{wei2022chain} demonstrate that chain-of-thought reasoning only reliably
emerges above a model-size threshold, suggesting that structured prompts may
impose undue cognitive overhead on smaller models---a pattern we observe in
our constrained tier.

\paragraph{Instruction following and format compliance.}
\citet{lou2023survey} survey instruction-following capabilities and find that
compliance degrades as instruction complexity increases, directly motivating
our harness-complexity investigation.
\citet{sclar2024quantifying} show that subtle changes in prompt formatting
cause up to 76-point performance differences across open-source models,
establishing that prompt structure is a major source of variance---a concern
we directly investigate at the harness level.
\citet{mizrahi2024state} argue that single-prompt evaluations are brittle
and call for multi-prompt evaluation, supporting our cross-harness design.
Work on structured output generation~\citep{geng2025jsonschemabench} shows that
JSON and schema compliance varies substantially across model families,
motivating our format-sensitive task category.
\citet{deng2025decoupling} find that separating task-solving from output formatting
improves both dimensions, consistent with the format violations we observe when
elaborate process instructions are mixed with output format requirements.

\paragraph{Agent scaffolding.}
ReAct~\citep{yao2022react} and Reflexion~\citep{shinn2023reflexion} demonstrate
that structured reasoning-action loops improve agent performance.
Our study complements this by asking whether different \emph{levels} of
harness structure suit different model tiers, and whether extended thinking
modes interact with harness structure in predictable ways.

\paragraph{Prompt complexity and performance inversion.}
\citet{schulhoff2024promptreport} provide a systematic survey of prompting
techniques and catalog conditions under which prompt engineering improves
or degrades performance, providing a broad empirical context for our
harness-complexity findings.
\citet{hakim2026brevity} find that brevity constraints on model outputs
reverse performance hierarchies across model scales---larger models become
relatively \emph{worse} when forced to be concise.
\citet{khan2025prompting} argue that increasing prompt specificity can
invert the expected performance ordering, with simpler prompts outperforming
engineered ones for capable models.
Both findings align with our harness-complexity paradox: the relationship
between instruction richness and task success is non-monotonic and
tier-dependent.

\paragraph{Self-correction and error recovery.}
\citet{li2025selfcorrection} decompose LLM self-correction and find an
accuracy--correction paradox where stronger models make ``deeper'' errors
that resist self-correction, while weaker models make more tractable surface
errors---a pattern partially consistent with our failure taxonomy.

\section{The HEAT-24 Benchmark}
\label{sec:benchmark}

\subsection{Workspace and Task Design}
\label{sec:tasks}

All tasks operate on a shared synthetic workspace containing twelve files:
configuration YAML and JSON, Python source and test files, Markdown
documentation, a CSV data file, and changelog fragments.
The workspace is initialized as a git repository before each run;
verification uses \texttt{git diff} to detect and scope file changes.
All twelve workspace files are injected into every harness prompt, enabling
models to read any file without external tool access.

We define 24 tasks across six categories (Table~\ref{tab:tasks}):

\begin{table}[ht]
\centering
\small
\begin{tabular}{lc}
\toprule
\textbf{Category} & \textbf{\# Tasks} \\
\midrule
\texttt{inspect\_local}         & 4 \\
\texttt{structured\_edit}       & 4 \\
\texttt{format\_sensitive}      & 4 \\
\texttt{verification\_recovery} & 4 \\
\texttt{repair}                 & 4 \\
\texttt{multi\_step\_ops}       & 4 \\
\bottomrule
\end{tabular}
\caption{HEAT-24 task categories (4 tasks each; 24 total).
\texttt{inspect\_local}: read files, return JSON;
\texttt{structured\_edit}: modify one file;
\texttt{format\_sensitive}: emit strict JSON schema;
\texttt{verification\_recovery}: fix bug, run tests;
\texttt{repair}: correct malformed content;
\texttt{multi\_step\_ops}: coordinate multiple files.
Each task has a deterministic binary verifier.}
\label{tab:tasks}
\end{table}

Tasks are designed to have deterministic, binary outcomes.
Verifiers check: JSON key presence and values, git-scoped file modifications,
YAML/JSON parse validity, and substring presence.
File modifications expressed in model output use a structured
\texttt{<<<WRITE:path>>>}$\cdots$\texttt{<<<END>>>} marker that the harness
runner parses and applies to the workspace before verification.
This marker format is included in the raw task instruction for all
file-modification tasks across all three harness conditions; harness
conditions differ only in the additional process instructions, scope
constraints, and verification specifications layered on top.

\subsection{Harness Conditions}
\label{sec:harness}

We define three harness conditions of increasing structural complexity:

\begin{description}
  \item[Light] A two-line prompt: role statement plus the raw task instruction.
        No format specification, no scope constraint, no verification procedure.
  \item[Balanced] Adds a four-step process template (plan, execute, check,
        respond) and lists the allowed files. No schema or verification spec.
  \item[Strict] Adds six explicit stages (preflight / plan / execute / verify /
        recover / report), an allowed-file list, explicit success criteria,
        a verification specification, and instructions to express file changes
        using the \texttt{<<<WRITE:path>>>} marker.
\end{description}

\subsection{Models}
\label{sec:models}

We evaluate six models spanning four capability tiers
(Table~\ref{tab:models}).
Tier assignments are based on deployment characteristics---parameter count
and inference infrastructure---\emph{not} on task performance, to avoid
circularity.
The goal is to test whether this conventional classification scheme predicts
harness sensitivity.
Each tier is represented by a single model in this study; tier-level claims
should therefore be interpreted as model-specific observations pending
replication with additional models within each tier.
API-hosted models (Gemini 2.5 Flash, Qwen3.5-122B, GPT-OSS-120B) were
queried with provider-default temperature and sampling settings at the time
of the experiment; exact parameter values are logged in the released runner scripts.
Ollama-hosted models used \texttt{think=False} to suppress chain-of-thought
tokens, with context length fixed at 4096 tokens via per-model Modelfiles.
Qwen3.5-122B's ``Frontier-Reasoning'' classification reflects extended thinking
being enabled as an inference configuration choice across all runs, not an
architectural property; results may differ if extended thinking were disabled.

\begin{table*}[t]
\centering
\small
\setlength{\tabcolsep}{5pt}
\begin{tabular}{lllll}
\toprule
\textbf{Model} & \textbf{Tier} & \textbf{Params} & \textbf{Provider} & \textbf{Notes} \\
\midrule
Gemini 2.5 Flash   & Frontier-Proprietary & ---             & Google AI Studio  & Proprietary API \\
Qwen3.5-122B-A10B  & Frontier-Reasoning   & 122B (10B act.) & vLLM (self-hosted)& MoE; extended thinking enabled (think-budget default) \\
GPT-OSS-120B       & Strong-Open          & 120B            & Groq Cloud        & \\
\midrule
Qwen3.5:2B         & Constrained          & 2B              & Ollama (local)    & 4096-token context Modelfile \\
LLaMA 3.2          & Constrained          & 3B              & Ollama (local)    & 4096-token context Modelfile \\
Gemma4:e2B         & Constrained          & 2B              & Ollama (local)    & 4096-token context Modelfile \\
\bottomrule
\end{tabular}
\caption{Models evaluated across four capability tiers.
Tier assignments are based on deployment characteristics (parameter count and inference infrastructure),
not on task performance.
Model names for constrained-tier models reflect Ollama tags for the corresponding
Google/Meta/Alibaba base models.}
\label{tab:models}
\end{table*}

\subsection{Metrics and Failure Taxonomy}
\label{sec:metrics}

We report two metrics per run:
\begin{itemize}
  \item \textbf{TSR} (Task Success Rate): binary pass/fail as determined by
        the workspace verifier.
  \item \textbf{VTSR} (Verified Task Success Rate): identical to TSR in this
        benchmark (all verifiers complete without infrastructure error);
        the distinction is maintained for deployments where verifier failures
        must be separated from model failures.
\end{itemize}

Failures are assigned one of six labels by an automated rule-based
classifier that inspects git diff output, JSON parse results, and
test execution logs; no manual labeling was performed:
\texttt{format\_violation} (output not parseable as required schema),
\texttt{wrong\_answer} (parseable but incorrect value),
\texttt{wrong\_file} (modified a file outside the allowed set),
\texttt{missing\_change} (no file modification detected),
\texttt{unrelated\_change} (modification to a correct file but wrong content),
\texttt{tests\_still\_fail} (code change did not fix targeted test failure).

\section{Results}
\label{sec:results}

\subsection{Overall Performance by Harness Condition}
\label{sec:results-overall}

Table~\ref{tab:main} reports mean VTSR across 24 tasks for each model--harness
combination (24 runs per cell).
Because each tier is represented by a single model, cross-tier comparisons
are exploratory; differences between tier rows reflect model-specific
observations rather than tier-level laws.

\begin{table}[ht]
\centering
\small
\setlength{\tabcolsep}{3pt}
\begin{tabular}{lcccc}
\toprule
\textbf{Model} & \textbf{Tier} & \textbf{Light} & \textbf{Balanced} & \textbf{Strict} \\
\midrule
Gemini 2.5 Flash
  & FP  & \best{95.8} & 58.3 & 66.7 \\
Qwen3.5-122B-A10B
  & FR  & 87.5 & 75.0 & \best{91.7} \\
GPT-OSS-120B
  & SO  & \best{95.8} & \best{95.8} & 87.5 \\
\midrule
Qwen3.5:2B   & C & 0.0  & \best{58.3} & 4.2  \\
LLaMA 3.2    & C & 16.7 & 4.2  & \best{20.8} \\
Gemma4:e2B   & C & \best{91.7} & \best{91.7} & \best{91.7} \\
\bottomrule
\end{tabular}
\caption{Mean VTSR (\%) by model and harness condition ($n=24$ per cell,
$k=1$ repeat).
\textbf{Bold} = best per row.
Tier codes: FP = Frontier-Proprietary, FR = Frontier-Reasoning,
SO = Strong-Open, C = Constrained.
GPT-OSS-120B strict T13--T24 was re-evaluated using a second API key
after the original run exhausted the 200\,k TPD rate limit;
the re-evaluation used identical prompt templates and provider-default
parameters, completed within the same calendar week as the original run.
Representative 95\% Wilson CIs (for $n=24$): 95.8\%\,$\to$\,[78.9,\,99.9],
75.0\%\,$\to$\,[53.3,\,90.2], 58.3\%\,$\to$\,[36.6,\,78.2],
0\%\,$\to$\,[0.0,\,14.2].
Results should be interpreted as preliminary evidence pending $k\geq3$
repetitions for statistical reliability.}
\label{tab:main}
\end{table}

\begin{figure}[t]
\centering
\includegraphics[width=\columnwidth]{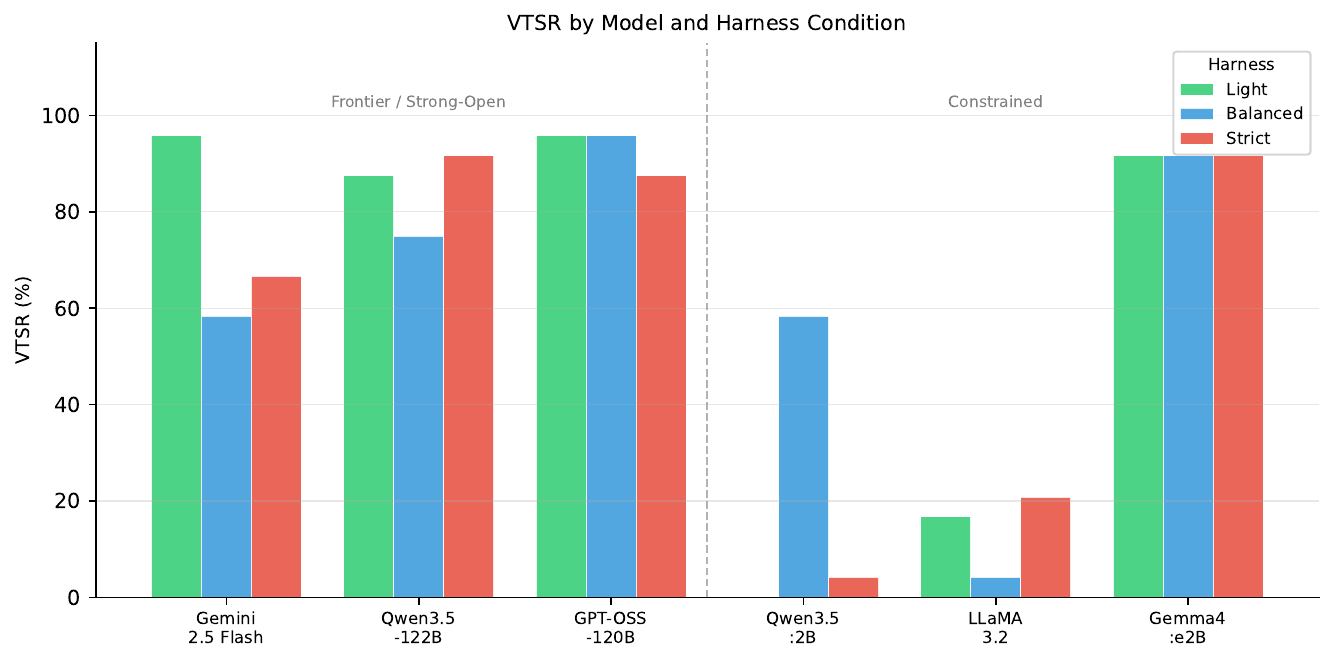}
\caption{VTSR (\%) by model and harness condition. The dashed line separates
frontier/strong-open (left) from constrained (right) tiers.}
\label{fig:vtsr}
\end{figure}

\subsection{Harness-Complexity Paradox: Frontier Chat Models}
\label{sec:paradox}

For Gemini 2.5 Flash (Frontier-Proprietary), light harness achieves
VTSR\,=\,\textbf{95.8\%}, dropping to 58.3\% under balanced ($-37.5$\,pp)
and 66.7\% under strict ($-29.2$\,pp).
The dominant failure mode in both complex conditions is
\texttt{format\_violation} (10 of 10 balanced failures; 8 of 8 strict failures).

Category-level analysis (Table~\ref{tab:category}) localizes the paradox to
\texttt{inspect\_local} and \texttt{format\_sensitive} tasks:
Gemini achieves 100\% on both categories under light harness but 0\% on
\texttt{inspect\_local} and 25\% on \texttt{format\_sensitive} under strict.
Tasks that require structured file editing (\texttt{structured\_edit},
\texttt{repair}) remain at 100\% across all harnesses, confirming that
the paradox is specific to \emph{format-sensitive output tasks}, not
general capability regression.

We hypothesize that elaborate multi-stage instructions shift the generation
distribution toward explanatory prose, causing the model to narrate its
reasoning rather than emitting the required JSON schema directly.
This aligns with the finding of \citet{deng2025decoupling} that task-solving and
output formatting compete as partially conflicting objectives.

\subsection{Non-Monotonic Pattern: Frontier Reasoning Models}
\label{sec:reasoning}

Qwen3.5-122B (Frontier-Reasoning, extended thinking enabled) exhibits
a \emph{non-monotonic} response to harness complexity: VTSR is lowest under
balanced harness (75.0\%), highest under strict (91.7\%), and intermediate
under light (87.5\%).
To our knowledge, this is the first empirical documentation of a
tier-specific non-monotonic interaction between harness complexity and model
type.

Notably, mean inference latency under strict harness (23.3\,s) is
\emph{lower} than under light (35.4\,s) or balanced (38.5\,s), consistent
with explicit constraints reducing the length of the model's thinking chains
before output.

Category analysis reveals structure: balanced-harness failures concentrate
in \texttt{inspect\_local} (only 25\% pass) and \texttt{format\_sensitive}
(50\% pass), both of which recover to 100\% under strict.
This pattern is consistent with the reasoning model using the strict
harness's explicit success criteria as scaffolding for its chain-of-thought,
reducing ambiguity in what constitutes a correct output.
Under the balanced harness, partial structure may create conflicting
signals that the extended thinking process amplifies rather than resolves.

\subsection{Strong-Open Model}
\label{sec:results-gpt}

GPT-OSS-120B (Strong-Open, via Groq) achieves equal and near-perfect
performance under light and balanced harnesses (95.8\% each), demonstrating
robustness to harness complexity across those two conditions.
Under strict harness, VTSR is 87.5\% (21/24), with T14, T16
(\texttt{wrong\_file}) and T24 (\texttt{format\_violation}) as the three
failures---a modest $-8.3$\,pp gap relative to light and balanced.
The initial strict run was contaminated by Groq's 200\,k tokens-per-day (TPD)
rate limit (T13--T24 returned empty outputs); the reported 87.5\% figure
is from a clean re-evaluation using a second API key.
This pattern---light $\approx$ balanced $\gg$ strict, with strict still
substantially above chance---distinguishes the Strong-Open tier from both
frontier tiers: it does not suffer the full harness-complexity paradox
(only $-8.3$\,pp, not $-29$\,pp), nor does it exhibit the non-monotonic pattern.

\subsection{Constrained-Tier Models}
\label{sec:results-constrained}

The three constrained-tier models exhibit three distinct patterns that together
reveal the heterogeneity within this tier.

\paragraph{Qwen3.5:2B---balanced-harness optimum.}
Qwen3.5:2B achieves 0\% VTSR under the light harness, 58.3\% under balanced,
and only 4.2\% under strict.
The light-harness failures are dominated by \texttt{wrong\_file} (15/24) and
\texttt{format\_violation} (9/24), indicating that without any structural
guidance the model cannot reliably locate the target file or produce required
output schemas.
The balanced harness's moderate structure (four-step process template plus
allowed-file list) provides sufficient scaffolding for this model to succeed
on the majority of tasks.
Strict harness then \emph{reverses} the gain: the six-stage process template
with verification specifications appears to exceed this model's
instruction-following capacity, collapsing performance back toward zero.
This \textbf{inverted-U pattern}---light $<$ strict $<$ balanced---stands in
direct contrast to both the frontier chat model (light $>$ strict $>$ balanced)
and the frontier reasoning model (strict $>$ light $>$ balanced), empirically
demonstrating that no harness condition is universally optimal.

\paragraph{LLaMA~3.2---low capability, harness-insensitive.}
LLaMA~3.2 achieves uniformly low VTSR across all conditions (light: 16.7\%,
balanced: 4.2\%, strict: 20.8\%).
Failures are dominated by \texttt{wrong\_file} across all harnesses, with
\texttt{format\_violation} rising under balanced and strict.
The near-flat performance curve ($\leq$21\% in any condition) indicates that
this model lacks the baseline instruction-following capability required to
benefit from structural harness guidance.
The slight strict advantage (20.8\% vs.\ 16.7\% light) is within noise and
does not constitute a reliable pattern.

\paragraph{Gemma4:e2B---frontier-level stability.}
Gemma4:e2B achieves 91.7\% VTSR under each of the three harness conditions,
matching the stability profile of GPT-OSS-120B despite having approximately
60$\times$ fewer parameters.
The two failures per condition span different failure types
(\texttt{wrong\_answer}, \texttt{format\_violation}, \texttt{wrong\_file})
rather than repeating the same label, indicating isolated variance rather than
systematic harness-induced failure.
This result challenges parameter count as a sufficient proxy for capability-tier
classification in harness-sensitivity studies: Gemma4:e2B's instruction-tuning
quality places its operational behavior firmly in the strong-open tier despite
its 2\,B parameter count.
We discuss this finding further in \S\ref{sec:discussion}.

\subsection{Performance by Task Category}
\label{sec:category}

Table~\ref{tab:category} reports macro-average VTSR by task category and
harness condition, separately for the frontier/strong tier (Gemini,
Qwen3.5-122B, GPT-OSS-120B) and the constrained tier (Qwen3.5:2B,
LLaMA~3.2, Gemma4:e2B).

\begin{table*}[t]
\centering
\small
\setlength{\tabcolsep}{7pt}
\begin{tabular}{lcccccc}
\toprule
 & \multicolumn{3}{c}{\textbf{Frontier/Strong}} & \multicolumn{3}{c}{\textbf{Constrained}} \\
\cmidrule(lr){2-4}\cmidrule(lr){5-7}
\textbf{Category} & \textbf{Li} & \textbf{Ba} & \textbf{St} & \textbf{Li} & \textbf{Ba} & \textbf{St} \\
\midrule
\texttt{inspect\_local}         & \best{100.0} & 41.7 & 66.7  & 41.7 & 50.0 & \best{58.3} \\
\texttt{structured\_edit}       & \best{100.0} & \best{100.0} & \best{100.0} & 33.3 & \best{50.0} & 41.7 \\
\texttt{format\_sensitive}      & \best{100.0} & 50.0 & 75.0  & \best{50.0} & 58.3 & 33.3 \\
\texttt{verification\_recovery} & 91.7 & \best{100.0} & 75.0  & 33.3 & \best{58.3} & 33.3 \\
\texttt{repair}                 & \best{100.0} & \best{100.0} & \best{100.0} & 33.3 & \best{50.0} & 41.7 \\
\texttt{multi\_step\_ops}       & 66.7 & 66.7 & \best{75.0}  & 25.0 & \best{41.7} & 25.0 \\
\bottomrule
\end{tabular}
\caption{Macro-average VTSR (\%) by category and harness ($n=12$ per cell for both tiers).
Li/Ba/St = Light/Balanced/Strict. \textbf{Bold} = best per tier$\times$category.}
\label{tab:category}
\end{table*}

\begin{figure}[t]
\centering
\includegraphics[width=\columnwidth]{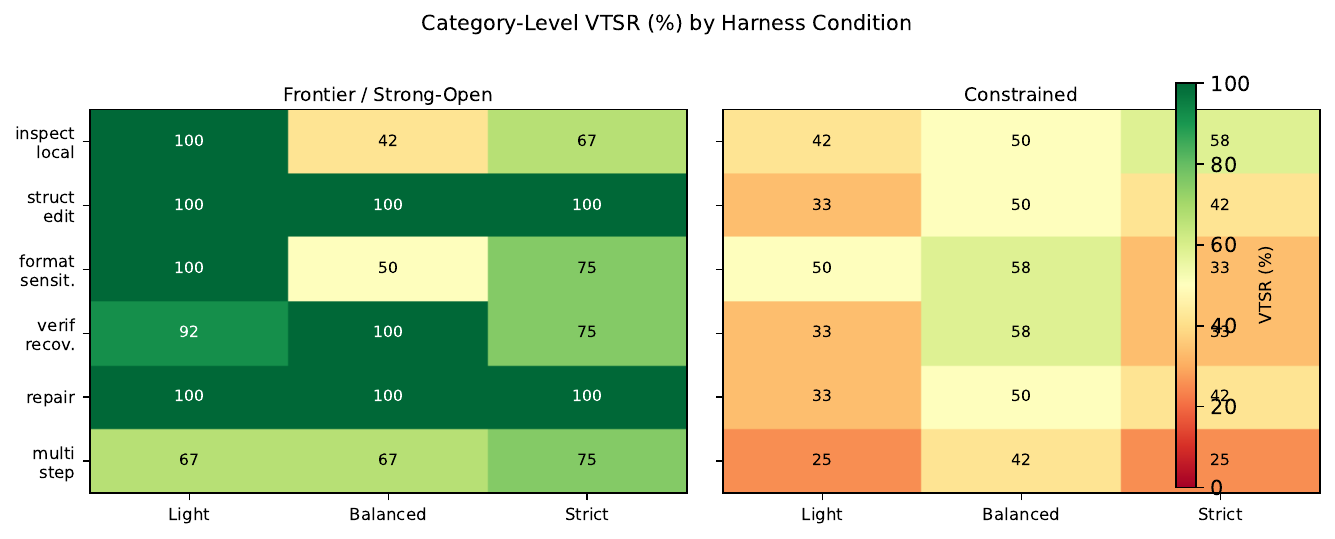}
\caption{Category-level VTSR heatmaps for frontier/strong-open (left)
and constrained (right) tiers. Green = high VTSR; red = low.}
\label{fig:category}
\end{figure}

For frontier/strong models, \texttt{structured\_edit} and \texttt{repair}
achieve 100\% across all harnesses, establishing them as reliable baselines
unaffected by harness complexity.
The \texttt{inspect\_local} and \texttt{format\_sensitive} categories show the
strongest harness sensitivity, confirming that JSON-output tasks are the
primary locus of format-violation failures.
\texttt{multi\_step\_ops} is the only category where strict (75.0\%) outperforms
both light and balanced (66.7\% each) for frontier/strong models.
For constrained models, balanced harness performs best or ties best in five
of six categories, with strict generally performing no better than light.
This pattern contrasts sharply with frontier chat models (where light dominates)
and reasoning models (where strict dominates), consistent with
the balanced harness providing just enough structural guidance without overloading
constrained models' instruction-following capacity.
The \texttt{multi\_step\_ops} category is universally the hardest for constrained
models (25\% light, 42\% balanced, 25\% strict), reflecting the inherent
difficulty of coordinating multiple file operations under limited capacity.

\subsection{Failure Label Distribution}
\label{sec:failures}

Table~\ref{tab:failures} reports the failure label counts by model and harness.

\begin{table}[ht]
\centering
\small
\setlength{\tabcolsep}{3pt}
\begin{tabular}{llcccccc}
\toprule
\textbf{Model} & \textbf{H}
  & \textbf{fv} & \textbf{wa} & \textbf{wf}
  & \textbf{mc} & \textbf{uc} & \textbf{tsf} \\
\midrule
Gemini 2.5 Flash  & Li & 1  & 0 & 0  & 0 & 0 & 0 \\
                  & Ba & 10 & 0 & 0  & 0 & 0 & 0 \\
                  & St & 8  & 0 & 0  & 0 & 0 & 0 \\
\midrule
Qwen3.5-122B      & Li & 2  & 0 & 0  & 0 & 0 & 1 \\
                  & Ba & 6  & 0 & 0  & 0 & 0 & 0 \\
                  & St & 1  & 0 & 0  & 0 & 0 & 1 \\
\midrule
GPT-OSS-120B      & Li & 1  & 0 & 0  & 0 & 0 & 0 \\
                  & Ba & 1  & 0 & 0  & 0 & 0 & 0 \\
                  & St & 1  & 0 & 2  & 0 & 0 & 0 \\
\midrule
Qwen3.5:2B        & Li & 9  & 0 & 15 & 0 & 0 & 0 \\
                  & Ba & 2  & 0 & 7  & 1 & 0 & 0 \\
                  & St & 8  & 0 & 15 & 0 & 0 & 0 \\
\midrule
LLaMA 3.2         & Li & 0  & 5 & 15 & 0 & 0 & 0 \\
                  & Ba & 9  & 0 & 14 & 0 & 0 & 0 \\
                  & St & 6  & 0 & 12 & 1 & 0 & 0 \\
\midrule
Gemma4:e2B        & Li & 1  & 1 & 0  & 0 & 0 & 0 \\
                  & Ba & 0  & 1 & 1  & 0 & 0 & 0 \\
                  & St & 1  & 0 & 1  & 0 & 0 & 0 \\
\bottomrule
\end{tabular}
\caption{Failure counts by model and harness.
H = Harness (Li/Ba/St).
fv = \texttt{format\_violation}, wa = \texttt{wrong\_answer},
wf = \texttt{wrong\_file}, mc = \texttt{missing\_change},
uc = \texttt{unrelated\_change}, tsf = \texttt{tests\_still\_fail}.
GPT-OSS-120B strict uses re-evaluated results (second API key).
\texttt{unrelated\_change} does not appear in any cell; it is retained
in the taxonomy for completeness.}
\label{tab:failures}
\end{table}

\begin{figure}[t]
\centering
\includegraphics[width=\columnwidth]{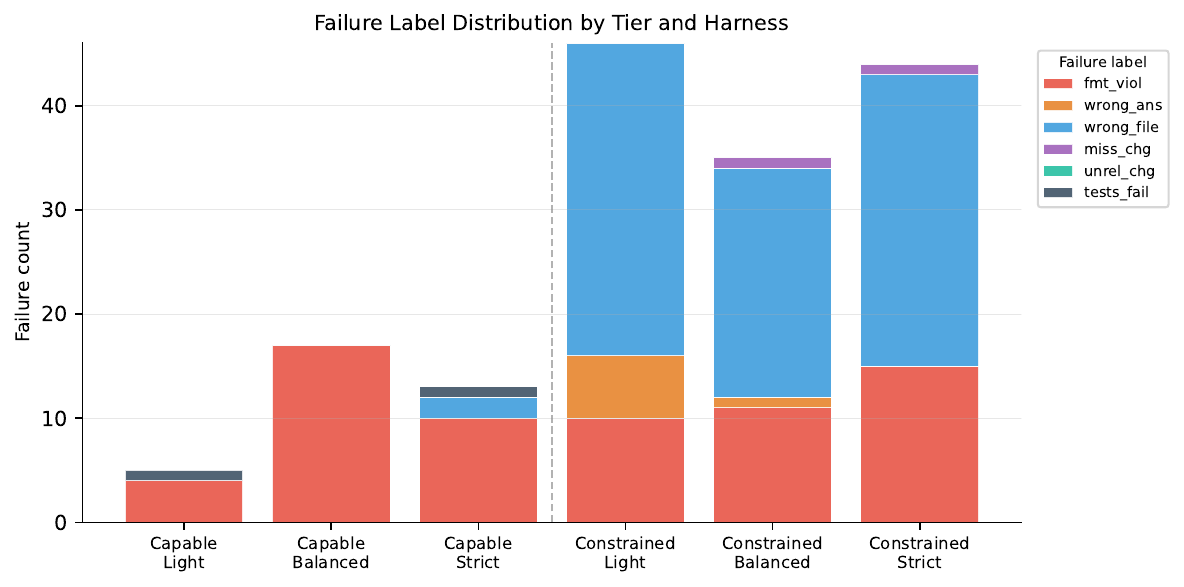}
\caption{Failure label distribution by tier and harness condition.
Capable models (top) are dominated by \texttt{format\_violation};
constrained models (bottom) by \texttt{wrong\_file}.}
\label{fig:failures}
\end{figure}

\texttt{format\_violation} is overwhelmingly the dominant failure mode introduced
by complex harnesses in capable models: 25 of 26 failures in the balanced and
strict conditions across Gemini 2.5 Flash and Qwen3.5-122B are format violations;
the single exception is one \texttt{tests\_still\_fail} in Qwen3.5-122B strict,
where the model understood the format but produced incorrect code changes.
No \texttt{wrong\_answer} or \texttt{missing\_change} failures appear in any
balanced or strict cell for these models.
This indicates that these models \emph{understand} the tasks but fail to
suppress explanatory prose when presented with process-heavy harness prompts.
GPT-OSS-120B strict failures (1 \texttt{format\_violation}, 2 \texttt{wrong\_file})
are distributed across three tasks (T14, T16, T24) without a clear categorical
pattern, consistent with model-level variance rather than systematic harness-induced
failure.
Constrained models (Qwen3.5:2B, LLaMA~3.2) exhibit a qualitatively different
failure signature: \texttt{wrong\_file} dominates under light harness, reflecting
a tendency to act on the wrong file without the structural guidance of an
allowed-file list.
LLaMA~3.2 alone shows five \texttt{wrong\_answer} failures under light harness,
the only tier$\times$harness cell where this label appears prominently,
suggesting limited task comprehension rather than format-compliance failure.

\subsection{Inference Latency}
\label{sec:latency}

Table~\ref{tab:latency} reports mean inference latency per task (seconds).

\begin{table}[h]
\centering
\small
\begin{tabular}{lccc}
\toprule
\textbf{Model} & \textbf{Light} & \textbf{Balanced} & \textbf{Strict} \\
\midrule
Gemini 2.5 Flash  & 2.6  & 3.8  & 6.0 \\
Qwen3.5-122B      & 35.4 & 38.5 & \best{23.3} \\
GPT-OSS-120B      & 8.4  & 8.5  & 7.0 \\
\midrule
Qwen3.5:2B        & 15.2 & 8.5  & 15.2 \\
LLaMA 3.2         & 1.3  & 2.7  & 2.9 \\
Gemma4:e2B        & 16.0 & 21.2 & 22.2 \\
\bottomrule
\end{tabular}
\caption{Mean inference latency (seconds per task). Gemini latency grows with
harness complexity. Qwen3.5-122B shows \emph{lower} latency under strict harness,
consistent with explicit constraints reducing thinking-chain length.
Constrained models (Ollama local) reflect hardware constraints; all values
include model-load time on shared GPU.}
\label{tab:latency}
\end{table}

Gemini's latency scales monotonically with harness complexity (2.6\,s $\to$
6.0\,s), consistent with longer prompts generating more verbose outputs.
Qwen3.5-122B shows the opposite trend: strict harness (23.3\,s) is 34\% faster than
light (35.4\,s), reinforcing the hypothesis that explicit constraints reduce
the length of the model's internal thinking chain.
GPT-OSS-120B latency is stable across conditions (7--9\,s),
suggesting it produces similarly sized outputs regardless of harness structure.
Among constrained models, LLaMA~3.2 is the fastest (1.3--2.9\,s) while
Gemma4:e2B is the slowest (16--22\,s); notably, Gemma4:e2B latency
\emph{increases} with harness complexity, mirroring Gemini's latency profile
and consistent with its similar frontier-level output quality.

\section{Discussion}
\label{sec:discussion}

\paragraph{Refuting the monotone inverse hypothesis.}
The central finding of this study is that harness sensitivity is
\emph{non-monotone} in capability tier.
The monotone inverse hypothesis predicts a single gradient: as capability
increases, optimal harness complexity decreases.
Our results break this gradient in two places simultaneously.
For the frontier \emph{chat} model (Gemini 2.5 Flash), strict harness
reduces VTSR by 29\,pp---consistent with the hypothesis direction, but
far larger in magnitude than expected.
For the frontier \emph{reasoning} model evaluated (Qwen3.5-122B), strict harness
\emph{increases} VTSR (+17\,pp over balanced) and \emph{reduces} latency
---directly contradicting the hypothesis, which predicts that a high-capability
model should need less harness structure.
The balanced harness occupies an awkward middle ground that helps neither
model type, suggesting that intermediate harness complexity is uniformly
suboptimal.

These results refute the monotone framing and instead suggest that model
\emph{type} (chat vs.\ reasoning) is an independent moderating variable
that capability tier alone cannot capture.
We note, however, that the chat/reasoning distinction emerged as a
\emph{post-hoc} interpretive frame from observing the results; it was not
a pre-specified moderating variable in the original study design.
Future work should treat this as an exploratory finding requiring
confirmatory replication with pre-registered hypotheses.

\paragraph{Tier-aware harness policy.}
The following guidelines reflect the specific model versions evaluated.
Because model families evolve rapidly, more durable guidance targets
model \emph{type} (chat vs.\ reasoning) and instruction-tuning quality
rather than specific model names; empirical re-evaluation is recommended
whenever models are updated or replaced.
Practitioners without benchmark access can approximate tier placement using
a short harness probe (e.g., a 4--6 task subset of HEAT-24 covering
\texttt{inspect\_local} and \texttt{format\_sensitive} categories), observing
whether format violations concentrate under complex or simple conditions.
Based on our results, we recommend:
\begin{itemize}
  \item \textbf{Frontier-Proprietary (chat)}: Use light harness for
        JSON-output and format-sensitive tasks; reserve strict harness for
        file-editing tasks where format compliance is not at risk.
        Light harness is also cost-optimal, as shorter prompts reduce
        API token consumption.
        Avoid balanced harness, which combines the complexity of strict
        with less structured guidance.
  \item \textbf{Frontier-Reasoning (extended thinking)}: Use strict harness
        across all task categories. Explicit success criteria and verification
        specifications align with the model's chain-of-thought and reduce
        both error rate and latency.
  \item \textbf{Strong-Open}: Light and balanced harnesses are equally
        effective (95.8\% each); strict harness incurs a modest $-8.3$\,pp
        penalty (87.5\%). For latency-sensitive deployments, light is preferred;
        for tasks requiring explicit constraints, strict remains viable.
  \item \textbf{Constrained (capable, e.g.\ Gemma4:e2B)}: Any harness works.
        This model's instruction-tuning quality renders it operationally
        equivalent to the strong-open tier; tier-aware routing based on
        parameter count alone would misclassify it.
  \item \textbf{Constrained (moderate, e.g.\ Qwen3.5:2B)}: Use balanced harness.
        The four-step process template and allowed-file list provide enough
        structural guidance to activate task understanding without overloading
        instruction-following capacity.
        Avoid light harness (catastrophic \texttt{wrong\_file} failures)
        and strict harness (instruction overload collapses performance).
  \item \textbf{Constrained (low capability, e.g.\ LLaMA~3.2)}: No harness
        condition yields reliable performance; deployment of this tier for
        workspace-editing tasks is not recommended without further capability
        improvement.
\end{itemize}

\paragraph{Instruction-tuning quality supersedes parameter count.}
Gemma4:e2B achieves 91.7\% VTSR at all harness conditions despite having
$\approx$2\,B parameters, matching the strong-open tier model (GPT-OSS-120B,
120\,B parameters) on stability and approaching the frontier chat model
(Gemini 2.5 Flash) on peak performance.
This result challenges the common practice of classifying models into
capability tiers by parameter count alone.
For harness-sensitivity prediction, a model's instruction-tuning quality---its
trained ability to comply with structured task directives and emit formatted
outputs---is a more reliable predictor than raw model scale.
Future tier-aware deployment frameworks should assess instruction-following
capability empirically (e.g., on a held-out harness probe) rather than relying
on parameter count as a proxy.
We also note an alternative interpretation: Gemma4:e2B's result may
indicate a tier-\emph{taxonomy} failure rather than a capability insight.
If this model genuinely belongs to the strong-open tier by instruction-tuning
quality, its placement in the constrained tier may inflate apparent
between-tier differences rather than demonstrating that parameter count is
a poor proxy.
Distinguishing these interpretations requires additional probing
beyond HEAT-24.

\paragraph{Format violations as systemic risk.}
Across all capable models, \texttt{format\_violation} is the dominant
harness-induced failure, never \texttt{wrong\_answer}.
This is a critical observation: the models are capable of solving the tasks
but fail operationally because they cannot resist injecting explanatory prose.
Future harness designs should separate the process-instruction component from
the output-format specification, presenting the latter immediately before the
model's generation window~\citep{deng2025decoupling}.
Alternatively, post-generation extraction (regex or schema-constrained
decoding) could recover JSON fields from verbose outputs.

\paragraph{Reasoning models and thinking-mode interaction.}
The strict harness interacts with extended thinking modes in a way
consistent with explicit constraints narrowing the reasoning model's
search space, producing shorter thinking chains and better outputs.
Whether this effect generalizes to other reasoning models or task domains
remains an open question requiring controlled study.

\paragraph{Limitations and threats to validity.}
\textit{External validity.}
Our workspace is synthetic; real-world repositories are larger, noisier,
and involve multi-file dependencies not present in our 12-file setup.
Results may not transfer directly to production software engineering tasks
such as those in SWE-bench~\citep{jimenez2024swebench}.

\textit{Statistical.}
The experiment uses a single repeat per condition ($k=1$); Wilson 95\%
confidence intervals for 24-task cells are wide (e.g., 58.3\%\,$\to$\,
$[36.6,78.2]$\%), so individual cells should be interpreted as
preliminary evidence.
Future work should add $k\geq3$ repetitions.

\textit{Internal validity.}
GPT-OSS-120B strict T13--T24 required re-evaluation with a second API key
after the original run hit Groq's 200\,k TPD rate limit; the re-run
showed T14 and T16 (\texttt{wrong\_file}) and T24 (\texttt{format\_violation})
failing, which may reflect model variance rather than harness effects.
Qwen3.5-122B was evaluated with extended thinking enabled across all harness
conditions; we cannot isolate the contribution of extended thinking from the
harness condition itself, so the non-monotonic pattern may reflect a
thinking-mode--harness interaction rather than a pure harness effect.
Constrained-tier models run with 4096-token context Modelfiles; tasks requiring
long outputs or large workspace context may be artificially penalised
relative to unconstrained inference, potentially contributing to the low
VTSR observed under some harness conditions.

\textit{Construct validity.}
Tier assignments are based on deployment characteristics (parameter count
and infrastructure) to avoid circularity; the finding that Gemma4:e2B
exceeds its assigned tier in performance is therefore a genuine empirical
result, not an artifact of classification.
Qwen3.5-122B is a Mixture-of-Experts model with $\approx$10B active
parameters; its ``Frontier-Reasoning'' classification reflects extended-thinking
capability, not parameter count.

\section{Conclusion}
\label{sec:conclusion}

The monotone inverse hypothesis---that higher-capability models need less
harness structure, forming a predictable gradient---does not hold.
Across 432 runs on HEAT-24, evidence suggests that harness sensitivity is
non-monotone across the models evaluated, and depends jointly on model type
(chat vs.\ reasoning) and instruction-tuning quality rather than capability
tier alone:
the evaluated frontier chat model benefits from light harness,
the evaluated frontier reasoning model benefits from strict harness, and a
2\,B constrained model matches strong-open stability regardless of harness
condition.
These results reject a single universal harness policy and call for
empirical tier-aware and type-aware harness selection.
HEAT-24, benchmark code, and full results will be released upon acceptance.

\section*{Acknowledgements}
Experiments were conducted using the Google AI Studio API,
Groq Cloud API (Groq, Inc.),
a self-hosted vLLM inference service,
and local Ollama inference.
We thank the providers of open-weight models evaluated in this work.

\bibliography{references}

\end{document}